\documentclass[runningheads]{llncs}
\usepackage[T1]{fontenc}
\usepackage{todonotes}
\usepackage{pifont}
\usepackage{graphicx}
\usepackage{amsmath}
\usepackage{amsfonts}
\usepackage{mathabx}
\usepackage{multirow}
\usepackage{xurl}

\usepackage{algorithm}
\usepackage{algpseudocode}

\usepackage{algorithm}
\usepackage{algpseudocode}
\algrenewcommand\algorithmicrequire{\textbf{Input:}}

\begin{document}
\title{Propagation of interval belief structures and imprecise copulas for neural network verification}
\titlerunning{Propagation of IBSs and imprecise copulas for NN verification}
%
%
\author{Francesc Pifarre-Esquerda\inst{*} \and
Eric Goubault\inst{*} \and Sylvie Putot\inst{*}}
\authorrunning{F. Pifarre-Esquerda et al.}
%
\institute{* LIX, CNRS, École polytechnique, Institut Polytechnique de Paris, Palaiseau, France}
\maketitle              
\begin{abstract}
Quantitative verification of neural networks requires reasoning about probabilities under substantial uncertainty in both input distributions and their dependence structure. In realistic settings, this information is often only partially specified, and assuming precise probabilistic models can lead to unreliable results.

We propose a sound framework for quantitative verification under imprecise probabilistic information, combining interval belief structures to represent marginal uncertainty with imprecise copulas to model uncertain dependence. 
We develop a propagation method for imprecisely coupled interval belief structures through feed-forward neural networks. Using mixed imprecise copula volumes, we derive sound push-forward constructions through affine transformations and activation functions. The resulting output can provide guaranteed lower and upper bounds on probabilistic safety properties, valid for all probability models compatible with the specified imprecise inputs.

\keywords{Imprecise probability \and Neural networks \and Interval belief structures \and Imprecise copulas \and Verification.}
\end{abstract}
\section{Introduction}

Neural networks (NNs) are increasingly deployed in safety-critical and high-risk domains, making the verification of their behaviour a central concern. Over the past years, significant progress has been made in the qualitative verification of NNs, where one seeks to determine whether a given safety specification is satisfied or violated for all admissible inputs. Sound and scalable techniques based on abstract interpretation have been developed \cite{singh2019abstract,stars19,prima}. 

By contrast, quantitative or probabilistic verification of neural networks taking probabilistic elements of the inputs into account remains comparatively underdeveloped. A central challenge in these quantitative approaches to neural network analysis is to represent both marginal uncertainty of each input feature and uncertainty about the dependence structure between features while still enabling sound forward propagation through affine layers and non-linear activations. However, in realistic applications, input uncertainty is rarely known precisely as marginal distributions may be only partially specified, and assumptions about dependence can be heuristic or unjustified. Collapsing epistemic uncertainty into a single precise probabilistic model can lead to a false sense of confidence \cite{ferson1996whereof,falseconfidence1} and remains a key obstacle to deploying trustworthy and interpretable NN models \cite{singh2025safety}.

This challenge is particularly relevant for probabilistic verification of NNs, where the objective is to bound the probability that a safety property holds under uncertain inputs. Existing approaches often rely on sampling-based methods \cite{wengPROVENCertifyingRobustness2018,balutaScalableQuantitativeVerification2021,pautovCCCERTProbabilisticApproach2022}, and therefore cannot provide guaranteed bounds. Recent guaranteed methods include probability stars \cite{tranQuantitativeVerificationNeural2023}, branch-and-bound algorithms \cite{boetiusSolvingProbabilisticVerification2024}, and Dempster-Shafer structures \cite{goubaultZonotopicDempsterShaferApproach2025a}.

Imprecise probability theory provides a principled mathematical framework to address these issues by explicitly representing both aleatoric uncertainty (intrinsic randomness) and epistemic uncertainty (lack of knowledge). Rather than committing to a single probability distribution, uncertainty is modelled by credal sets consistent with available information. This approach has proved valuable in a variety of fields, including engineering analysis \cite{beerImpreciseProbabilitiesEngineering2013} or dynamical systems \cite{grayVerifiedPropagationImprecise2024}. 

The central contribution of this work is the application of imprecise probability tools to the verification of NNs. We develop a sound propagation method for marginal interval belief structures joined by an imprecise copula through feed-forward neural networks. The propagation relies on imprecise copula volumes evaluated on conservative/optimistic quantile rectangles, yielding computable lower and upper bounds on the probability mass assigned to each input cell. We then derive push-forward constructions through affine transformations and activation functions to obtain both output marginal IBSs and an output imprecise copula that captures both the specified dependencies between inputs and dependencies induced by the network itself. This enables sound evaluation of probabilistic safety properties at the network output.

\section{Imprecise model: marginal interval belief structures and imprecise copulas}
\subsection{Interval belief structures}
We use as basic imprecise probability structure the probability box \cite{Ferson2003}, usually referred to as p-box in short. P-boxes represent the credal sets (sets of probability distributions) that lie between a pair of bounding or envelope CDFs. The probability distributions within the credal set capture the aleatoric element of uncertainty while the set itself represents the lack of full epistemic knowledge.

Performing operations on p-boxes can be done through convolutions \cite{williamsonProbabilisticArithmeticNumerical1990}, but analytical solutions to these are not generally available. To facilitate operations on p-boxes, Interval Dempster-Shafer structures (DSIs) are commonly used \cite{yagerDempsterShaferBeliefStructures2001,Ferson2003}. DSIs offer discrete outer approximations of p-boxes and provide a way to perform sound approximations of arithmetic operations on them. They have also been applied to NN contexts in \cite{goubaultZonotopicDempsterShaferApproach2025a}.

Interval Belief Structures (IBSs) further generalize these classical interval Dempster-Shafer structures by allowing the basic probability masses attached to focal sets to be imprecise and given in terms of intervals. They were introduced simultaneously as generalization of DSIs in \cite{yagerDempsterShaferBeliefStructures2001} and focusing on their induced belief functions in \cite{denoeuxReasoningImpreciseBelief1999}. IBSs have previously been successfully applied both to general evidential reasoning \cite{wangEvidentialReasoningApproach2006} and to domain specific tasks such as diabetes diagnostics \cite{sevastianovFrameworkRulebaseEvidential2012}. In our propagation through neural networks, IBSs will be used as the basic unit of computation for marginal distributions.

\begin{definition}[Interval belief structure]
\label{ibs}
Let $\mathcal{F}=\{F_1,\dots,F_N\}$ be a family of non-empty subsets of $\mathcal{X}$, referred to as focal sets.  
An interval belief structure on $\mathcal{X}$ is specified by a collection of intervals
\[
I = \{ \langle F_i,\,[\underline{m}_i,\overline{m}_i]\rangle \mid i=1,\dots,N\},
\]
with $0 \le \underline{m}_i \le \overline{m}_i \le 1$ for all $i$, together with the requirement that there exists at least one basic probability assignment $m$ such that
\[
m(F_i) \in [\underline{m}_i,\overline{m}_i], \quad i=1,\dots,N,
\qquad
\sum_{i=1}^N m(F_i) = 1,
\]
which means that the IBS is not empty.  

The set of all probability assignments compatible with the IBS $I$ is the credal set of beliefs
\[
\mathcal{M}(I) 
= \bigl\{ m \mid m(F_i)\in[\underline{m}_i,\overline{m}_i], \ \sum_{i=1}^N m(F_i)=1 \bigr\}.
\]
\end{definition}

For an IBS to be well-defined, it needs to be non-empty in the credal set sense, meaning that its bounds must allow for at least one valid distribution. This non-emptiness property can be characterized as follows:

\begin{proposition}[Non-emptiness of IBS]
A necessary and sufficient condition for the existence of at least one compatible $m$ (non-emptiness of $\mathcal{M}(I)$) is
$\sum_{i=1}^N \underline{m}_i \,\le\, 1 \,\le\, \sum_{i=1}^N \overline{m}_i$
as given in Proposition 1 of \cite{denoeuxReasoningImpreciseBelief1999}.
\end{proposition}

Moreover, the idea of the normalization of IBS is also relevant. This relates to the tightness of the mass bounds with respect to their associated credal set. If an IBS is normalized, the mass bounds can be realized by a particular mass assignment or DSI within the credal set. We will aim to work with normalized IBSs for increased tightness and well-defined quantile grids in relevant calculations.

\begin{definition}[Normalized interval belief structure]
\label{normalized-ibs}
Let $I = \{ \langle F_i,\,[\underline{m}_i,\overline{m}_i]\rangle \mid i=1,\dots,N\}$ be an interval belief structure with $0 \le \underline{m}_i \le \overline{m}_i \le 1$ for all $i$, and assume it is non-empty so that $\mathcal{M}(I)\neq\emptyset$.

We say that $I$ is a normalized interval belief structure (in the sense of Wang and Elhag \cite{wangCombinationNormalizationIntervalvalued2007,sevastianovFrameworkRulebaseEvidential2012}) if:
\[
\sum_{j=1}^N \overline{m}_j - (\overline{m}_i - \underline{m}_i) \ge 1
\quad\text{and}\quad
\sum_{j=1}^N \underline{m}_j + (\overline{m}_i - \underline{m}_i) \le 1,
\qquad \forall i \in \{1,\dots,n\}.
\]
Equivalently, for every $i$ the whole interval $[\underline{m}_i,\overline{m}_i]$ consists of admissible values $m(F_i)$ for some $m \in \mathcal{M}(I)$. In this sense normalized interval belief structures form a tight representation of their associated credal set. This condition is also equivalent to: 
$\underline{m}_i = \inf_{m \in \mathcal{M}(I)} m(F_i)$ and
$\overline{m}_i = \sup_{m \in \mathcal{M}(I)} m(F_i)$.
\end{definition}

\begin{proposition}[IBS normalization]
\label{ibs-normalization}
Let $I = \{ \langle F_i,\,[\underline{m}_i,\overline{m}_i]\rangle \mid i=1,\dots,N\}$ be an IBS. It can be normalized by retaking the mass bounds at each focal element $i$ such that the bounds are given by: \[
\max\left[ \underline{m}_i,\; 1 - \sum_{j \ne i} \overline{m}_j \right]
\;\le\;
m(F_i)
\;\le\;
\min\left[ \overline{m}_i,\; 1 - \sum_{j \ne i} \underline{m}_j \right],
\qquad i = 1,\ldots,N.
\]

This construction has been adapted from Section 4 of \cite{wangCombinationNormalizationIntervalvalued2007}.
\end{proposition}

In the semantics of probability distributions, IBSs induce belief functions on their elements. We restrict ourselves here to the unidimensional case where $\mathcal{X}=\mathbb{R}$ and all focal sets $F_i$ are intervals. IBSs represent univariate distributions and their associated uncertainty.  IBSs become DSIs in the "degenerate" or limit case where the upper and lower mass bounds coincide.

A p-box can be recovered from an IBS by taking the broadest envelopes possible. By doing so, a lot of extra information contained by the IBS on the internal structure of the resulting p-box credal set is lost, but the p-box credal set is an outer approximation of the original IBS.

\begin{definition}[IBS belief, plausibility, and induced p-box]
\label{ibs-bel-pbox}
Let $I$ be an IBS with credal set $\mathcal M(I)$. For any $m\in\mathcal M(I)$ and $A\subseteq X$, we define
$\text{Bel}_m(A)=\sum_{F_i\subseteq A} m(F_i)$,
$\text{Pl}_m(A)=\sum_{F_i\cap A\neq\emptyset} m(F_i)$
which induce bounds: 
\[
\underline{\text{Bel}}(A)=\inf_{m\in\mathcal M(I)} \text{Bel}_m(A), \quad
\overline{\text{Pl}}(A)=\sup_{m\in\mathcal M(I)} \text{Pl}_m(A).
\]

\noindent These define the IBS's CDF envelopes: 
\[
\underline F(x):=\underline{\text{Bel}}((-\infty,x]), \qquad
\overline F(x):=\overline{\text{Pl}}((-\infty,x]).
\]
It holds that $\underline F\le\overline F$, both are non-decreasing, and $[\underline F,\overline F]$ is a p-box representing all distributions compatible with $I$.
\end{definition}

To relate IBSs to copulas, we will use the quantile levels. These lie in the CDF codomain as the accumulated probability in each focal element. For IBSs, these quantile levels are imprecise and given only within intervals.

\begin{proposition}[Quantile levels for IBS]
\label{quantile-levels-ibs}
Let $I$ be a normalized IBS. We define its quantile levels as given by its lower belief and upper plausibility:
\[
\underline{\alpha}_i = \max \left \{ \sum_{k=1}^i \underline{m}_k \, , \, 1 - \sum_{k=i+1}^N \overline{m}_k\right \}
\qquad
\overline{\alpha}_i = \min \left \{ \sum_{k=1}^i \overline{m}_k \, , \, 1 - \sum_{k=i+1}^N \underline{m}_k\right \}.
\]
We use the boundary conventions $\underline{\alpha}_0=\overline{\alpha}_0=0$ and $\underline{\alpha}_N=\overline{\alpha}_N=1$.

Since the upper and lower masses need not add to $1$, the second argument on the min/max assigns the minimum mass to the first $i$ elements, unless the maximum/minimum mass of the remaining elements is not enough/too much to reach $1$ for the last quantile level. This construction follows Definition 3 of \cite{wangCombinationNormalizationIntervalvalued2007}. 
\end{proposition}

We refer to the interval quantile bounds $[\underline{\alpha}_i, \overline{\alpha}_i]$ jointly as $\alpha_i$. For quantile levels and IBSs indices to be well-defined we require a total order between the focal elements. By convention when working with p-boxes, we will take the lexicographical order on the pair given by the focal interval bounds \cite{Ferson2003}. 

\subsection{Imprecise copulas}
The same arguments that justify the employment of imprecise probability structures to model marginal distributions can also be raised for imprecise dependence structures.
Copulas allow to separate the information in the joint dependence of random variables and their marginal distribution through Sklar's seminal theorem \cite{nelsenIntroductionCopulas2010}. Imprecise copulas allow us to do the same while accounting for uncertainty in their dependence.

The main result justifying our decision to employ IBS marginals joined by an imprecise copula, which we will refer to jointly as (imprecisely) coupled IBSs, is the extension of Sklar's theorem to the imprecise domain. Coherent imprecise copulas allow to relate imprecise marginals to imprecise multivariate distributions with results following from Sklar's seminal theorem for "precise" copulas \cite{montesSklarsTheoremImprecise2015,omladicFullScaleSklars2020,omladicMultivariateImpreciseSklar2022}. In what follows, when we speak of coupled IBSs we mean a multivariate distribution such that each marginal basic probability assignment lies in the IBSs marginal credal sets and the dependence is described by some unknown copula $C$ lying between the quasi-copula envelopes $\underline Q\le C\le \overline Q$.

This section introduces necessary structures and results to justify this modelling decision, and lays down the groundwork to later define the propagation of coupled IBSs through a NN. The basic building blocks of imprecise copulas are not properly copulas, but a relaxed version of them called quasi-copulas.
\begin{definition}[Quasi-copulas]
For any point $(u_1, \ldots, u_{j-1}, u_{j+1}, \ldots, u_n) \in \mathbb{I}^{n-1}$ and any $t \in \mathbb{I}$ (respectively $\overline{\mathbb{I}}$). A $n$-dimensional quasi-copula $Q$ is a function $Q: \mathbb{I}^n \to \mathbb{I}$ that satisfies the following conditions:

\begin{enumerate}
    \item[(i)] For every $j = 1, \dots, n$ we have $Q(1, \ldots, 1, u_j, 1, \ldots, 1) = u_j$.
    
    \item[(ii)] $Q$ is increasing in each of its variables, i.e., for every $j = 1, \dots, n$ and for every point $(u_1, \ldots, u_n) \in \mathbb{I}^{n}$ the value $Q(\mathbf{u})$ 
    is non-decreasing as any $u_j$ increases with the rest remaining fixed. 
    \item[(iii)] $Q$ is Lipschitz, i.e., if $\mathbf{u}, \mathbf{v} \in \mathbb{I}^n$, then
    $|Q(\mathbf{v}) - Q(\mathbf{u})| \leq \sum_{j=1}^n |v_j - u_j|$.
    \item[(iv)] for every $j = 1, \dots, n$ and for every point $(u_1, \ldots, u_{j-1}, u_{j+1}, \ldots, u_n) \in \mathbb{I}^{n-1}$ we have
    $Q(u_1, \ldots, u_{j-1}, 0, u_{j+1}, \ldots, u_n) = 0$.
\end{enumerate}
\end{definition}

Quasi-copulas give the envelopes of the credal set which defines an imprecise copula. To have an imprecise copula, however, care is needed to ensure that they are well-defined. This refers to issues of non-emptiness and coherence, which remained open for some time after the original introduction of imprecise copulas in \cite{montesSklarsTheoremImprecise2015}. In essence, the original characterization of imprecise copulas given in \cite{montesSklarsTheoremImprecise2015} was shown to be flawed in that it allowed for imprecise copulas which had no copula within (counterexample given in \cite{omladicFinalSolutionProblem2020}). Therefore we follow the tighter characterization of a well-defined imprecise copula of \cite{omladicFinalSolutionProblem2020} and its generalization by the same authors to higher-dimensional cases in \cite{omladicMultivariateImpreciseSklar2022}.

\begin{definition}[Avoiding sure loss or non-emptiness]
We consider a pair of quasi-copulas $(\underline{Q}, \overline{Q})$ to avoid sure loss (or to be non-empty in the credal set sense) if there exists a proper copula $C$ with $\underline{Q} \le C \le \overline{Q}$.
\end{definition}

\begin{definition}[Coherent imprecise copulas]
We will call a pair $[\underline{Q}, \overline{Q}]$ of $n$-variate quasi-copulas a coherent imprecise copula if they are non-empty (avoid sure loss) and:
$\underline{Q} = \inf \{C \mid \underline{Q} \le C \le \overline{Q} \}$, and 
$\overline{Q} = \sup \{C \mid \underline{Q} \le C \le \overline{Q} \}$.

The infimum and supremum are taken pointwise with respect to the product order. This definition follows the characterization of multivariate imprecise copulas introduced after Th. 16 of \cite{omladicMultivariateImpreciseSklar2022}. We will call the quasi-copulas the bounds or envelopes of the imprecise copula.
\end{definition}

We point out the similarity of the notion of a coherent imprecise copula to the normalization of an IBS: both require the envelopes describing the credal sets (either masses on IBS or quasi-copulas in the imprecise copula) to be tight. The volumes of imprecise copula, which will be used for propagation, generalize the precise copula volumes which are the measure induced by copulas on rectangles. For an imprecise copula, we can derive two dual induced volumes:

\begin{definition}[Mixed imprecise copula volume]
Let $[\underline{Q}, \overline{Q}]$ be an imprecise copula and let 
\[
    P \;=\; [\underline{p}_1,\overline{p}_1]\times\cdots\times[\underline{p}_n,\overline{p}_n]
    \;\subseteq\; [0,1]^n
\]
be a rectangle. We will refer to as the lower and upper mixed imprecise volumes of $P$, denoted $\underline{V}_{[\underline{Q}, \overline{Q}]}(P)$ and $\overline{V}_{[\underline{Q}, \overline{Q}]}(P)$ as:
\begin{align*}
    & \underline{V}_{[\underline{Q}, \overline{Q}]}(P) =
    \sum_{\begin{subarray}{c}
   \mathbf{p}\in\mathrm{vertices}(P) \\
   \operatorname{sign}(\mathbf{p}) > 0
    \end{subarray}}
    \operatorname{sign}(\mathbf{p})\, \underline{Q}(\mathbf{p}) +     \sum_{\begin{subarray}{c}
   \mathbf{p}\in\mathrm{vertices}(P) \\
   \operatorname{sign}(\mathbf{p}) < 0
    \end{subarray}}
    \operatorname{sign}(\mathbf{p})\, \overline{Q}(\mathbf{p}) \\
    & \overline{V}_{[\underline{Q}, \overline{Q}]}(P) =
    \sum_{\begin{subarray}{c}
   \mathbf{p}\in\mathrm{vertices}(P) \\
   \operatorname{sign}(\mathbf{p}) > 0
    \end{subarray}}
    \operatorname{sign}(\mathbf{p})\, \overline{Q}(\mathbf{p}) +    \sum_{\begin{subarray}{c}
   \mathbf{p}\in\mathrm{vertices}(P) \\
   \operatorname{sign}(\mathbf{p}) < 0
    \end{subarray}}
    \operatorname{sign}(\mathbf{p})\, \underline{Q}(\mathbf{p})
\end{align*}

where the sum is over all vertices $\mathbf{p}=(p_1,\ldots,p_n)$ with $p_i\in\{\underline{p}_i,\overline{p}_i\}$, and
\[
    \operatorname{sign}(\mathbf{p}) =
    \begin{cases}
        +1, & \text{if $\mathbf{p}$ has an even number of lower bounds } \underline{p}_i,\\[2pt]
        -1, & \text{if $\mathbf{p}$ has an odd number of lower bounds } \underline{p}_i.
    \end{cases}
\]
\end{definition}

These mixed volumes are based on the $L$ operator of \cite{omladicFinalSolutionProblem2020}, which corresponds to our $\overline{V}_{[\underline{Q}, \overline{Q}]}$, and $\underline{V}_{[\underline{Q}, \overline{Q}]}$ is constructed as its dual counterpart. As $\underline{Q}=\overline{Q}=C$, we have a precise copula (if the more restrictive copula axioms are satisfied for $C$). Then $\underline{V}_{[\underline{Q}, \overline{Q}]}(P)=\overline{V}_{[\underline{Q}, \overline{Q}]}(P)=V_C(P)$ gives the standard copula volume. For quasi-copula bounds, $\underline{V}_{[\underline{Q}, \overline{Q}]}(P)$ may be negative as quasi-copulas need not be $n$-increasing. Also note that while the usual copula volume is a measure and additive with respect to disjoint unions of rectangles, the mixed imprecise copula volumes are not. Neither of the mixed volumes is a proper measure and $\overline{V}_{[\underline{Q}, \overline{Q}]}$ is only super-additive while $\underline{V}_{[\underline{Q}, \overline{Q}]}$ is sub-additive.

Mixed volumes will be key in our arithmetic to propagate coupled IBSs, but are also useful in the characterizations of the non-emptiness, and therefore the coherence, of an imprecise copula, by Theorem \ref{non-empty-positive-volumes}.

\begin{proposition}[Mixed volumes order]
Let $[\underline{Q}, \overline{Q}]$ be an imprecise copula and $C \in [\underline{Q}, \overline{Q}]$. For any rectangle $R$:
$\underline{V}_{[\underline{Q}, \overline{Q}]}(R) \le V_C(R) \le \overline{V}_{[\underline{Q}, \overline{Q}]}(R)$.
\end{proposition}

\begin{theorem}[Characterization of non-emptiness]
\label{non-empty-positive-volumes}
Let $\underline{Q} \le \overline{Q}$ be (discrete) quasi-copulas defined on $D \subseteq \mathbb{I}^n$. Then, there exists a copula $C$ defined on $D$ with
$\underline{Q} \le C \le \overline{Q}$ if and only if
   $\overline{V}_{[\underline{Q}, \overline{Q}]} \ge 0$
for all $R \in \mathcal{R}_D$ (set of all finite disjoint unions of rectangles with vertices on $D$). This result corresponds to Theorem 15 of \cite{omladicMultivariateImpreciseSklar2022} adapted to our notation.
\end{theorem}

\section{Propagation through neural networks}
In this section, we propagate IBSs and imprecise copulas through feed-forward neural networks in such a way that they over-approximate the distributions of the random variables in each layer. Two operations are composed to define an arbitrary feed-forward neural network: affine steps and activation functions.

\begin{definition}[Feed-forward Neural Network] We consider an $L-$layer fully-connected feed-forward neural network, for $L\in\mathbb N$ and $h_0,\dots,$ $h_L\in\mathbb N$ the layer widths, specified by 
$\{(W^k,b^k,\sigma^k)\}_{k=0}^{L-1}$, 
where for each layer $k=0,\dots,L-1$,
$W^k\in\mathbb R^{h_{k+1}\times h_k}$, 
and $\sigma^k: \ \mathbb R^{h_{k+1}}\to\mathbb R^{h_{k+1}}$
is an activation map applied coordinate-wise:
$\sigma^k(s)=\big(\sigma^k_1(s_1),\dots,\sigma^k_{h_{k+1}}(s_{h_{k+1}})\big)$ for 
$s\in\mathbb R^{h_{k+1}}$.

Given an input $x=x^0\in\mathbb R^{h_0}$, the network output $f_\theta(x)\in\mathbb R^{h_L}$ is defined recursively as:
$s^{k+1}=W^k x^k+b^k$, $x^{k+1}=\sigma^k(s^{k+1})$, 
for $k=0,\dots,L-1$, 
and $f_\theta(x):=x^L$. We denote the corresponding random vectors in the propagation as:
$\mathbf X^0:=\mathbf X$, 
$\mathbf{S}^{k+1}:=W^k\mathbf X^k+b^k$, 
$\mathbf{X}^{k+1}:=\sigma^k(\mathbf{S}^{k+1})$.
\end{definition}

First, we will define how to soundly apply the operation, affine or activation, on our marginal distributions represented by the IBSs. Secondly, we develop push-forward constructions to propagate the imprecise copulas through the same operations. We refer to this propagation method as a push-forward as the constructions propagate the quasi-copula envelopes through their induced volume on select rectangles defined on the IBS's quantile levels.

\subsection{Affine step}
Let $(X_i)_{i=1}^n$ be represented by normalized IBSs $I_{X_i} = \{ \mathbf{x}_{j_i}^i, \, \mathbf{m}_{j_i}^i\}$, where $j_i$ will be used as a running index iterating over the focal elements of the different marginals $X_i$. Also let $(X_i)_{i=1}^n$ be coupled by the multivariate imprecise copula with envelopes $[\underline{Q}, \overline{Q}]$ and with associated quantile levels $\left [\underline{\alpha}^i_{j_i}, \overline{\alpha}^i_{j_i} \right ]$. We define two possible rectangles upon which we will calculate the mixed copula volume: a lower and upper one:
\[
\underline{R}_{j_1, \dots, j_n} = \bigtimes_{i=2}^n [\overline{\alpha}^i_{j_i-1}, \underline{\alpha}^i_{j_i}]  \quad \text{and } \quad \overline{R}_{j_1, \dots, j_n} = \bigtimes_{i=2}^n  [\underline{\alpha}^i_{j_i-1}, \overline{\alpha}^i_{j_i}]
\]

Note that $\underline{R}{j_1, \dots, j_n}$ uses the highest possible previous quantile ($\overline{\alpha}^i_{j_i-1}$) and the lowest possible current quantile ($\underline{\alpha}^i_{j_i}$), creating the smallest possible quantile rectangle (empty if $\overline{\alpha}^i_{j_i-1} > \underline{\alpha}^i_{j_i}$) that could represent this focal element's probability mass in the copula space. This ensures we never overestimate the lower bound.
Analogously, $\overline{R}_{j_1, \dots, j_n}$ uses the lowest possible previous quantile ($\underline{\alpha}^i_{j_i-1}$) and the highest possible current quantile ($\overline{\alpha}^i_{j_i}$) to get the largest possible quantile rectangle so that we never underestimate the upper bound.

\begin{proposition}[Addition of imprecisely coupled IBSs]
\label{sum-ibs}
Let $Z = \sum_{i=1}^n X_i$. Then $I_Z = \{ \langle \mathbf{z}_{j_1,\dots,j_n}, \mathbf{r}_{j_1,\dots,j_n} \rangle\} $ such that for all $j_i = 1, \dots, N_i$ for $i = 1, \dots, n$ we have
\[
\mathbf{z}_{j_1,\dots,j_n} = \mathbf{x}_{j_1}^1 + \dots + \mathbf{x}_{j_n}^n
\quad \text{and} \quad
\mathbf{r}_{j_1,\dots,j_n} = \left [ \underline{V}_{[\underline{Q}, \overline{Q}]}(\underline{R}_{j_1, \dots, j_n}), \overline{V}_{[\underline{Q}, \overline{Q}]}(\overline{R}_{j_1, \dots, j_n}) \right ].
\]
\end{proposition}

For the focal elements, we use traditional interval arithmetic. For the upper bound, if the imprecise copula is coherent (or only non-empty), the upper mixed volume is guaranteed to be non-negative (Th. \ref{non-empty-positive-volumes}). We also note that the number of focal elements explodes after the affine transformation, with a new element for each input-cell combination $(j_1,\dots,j_n)$.

This IBS sum construction gives an outer approximation of the operation on its credal set. This quality holds as for $\underline{R}_{j_1, \dots, j_n} \subset R \subset \overline{R}_{j_1, \dots, j_n}$ a rectangle and $C \in [\underline{Q}, \overline{Q}]$ a copula we have the inequalities $V_C(R) \le V_C(\overline{R}) \le \overline{V}_{[\underline{Q},\overline{Q}]}(\overline{R})$.

Note that, however, the resulting IBSs may not be tight, as the proposed affine propagation does not preserve normalization. For further propagation, since we will need quantile levels which are only well-defined for normalized IBSs (Prop. \ref{quantile-levels-ibs}) the IBS is normalized by applying Proposition \ref{ibs-normalization}.

We also need to propagate the imprecise copula through the affine operation to obtain the dependence between the new variables created by the affine layer. Let $\mathbf{S} = (S_1, \dots, S_m)$ be the output variables defined by the linear sums $S_k = \sum_{i \in I_k} a_iX_i$ with $a_i$ scalars.
The input space is discretized by the focal elements of $\mathbf{X}$. For every multi-index $\mathbf{J} = (j_1, \dots, j_n)$, we determine the mass interval $[\underline{m}_{\mathbf{J}}, \overline{m}_{\mathbf{J}}]$ associated with the input cell $R_{\mathbf{J}}$. The focal elements and their mass bounds are computed using the mixed volumes of the input imprecise copula on the conservative and optimistic quantile rectangles $\underline{R}_{\mathbf{J}}, \overline{R}_{\mathbf{J}}$ as given in Prop. \ref{sum-ibs}.

Let $I_{S_k}$ be the resulting marginal IBS for each sum variable $S_k$, consisting of focal intervals $\{\mathbf{s}^k_{\ell}\}$ obtained by applying Proposition \ref{sum-ibs} and normalized by Prop. \ref{ibs-normalization}, and their associated quantile levels given by $\left \{ \{\gamma^k_{j_k}\}_{j_k=1}^{N_k} \right \}_{k=1}^m$. There exists a deterministic index mapping $\mathbf{L}^{\text{aff}}(\mathbf{J}) = (L^{\text{aff}}_1(\mathbf{J}), \dots, L^{\text{aff}}_m(\mathbf{J}))$ defined by the affine transformation such that the image of any input rectangle under $T$ is strictly contained within a specific output cell by tracking which focal elements of the original IBSs are added to form a new focal element on the output IBSs. 

The $\mathbf{L}^{\text{aff}}$ memory map is defined such that the indices of the focal elements of the variable resulting from the affine transform $I_Z = \{ \langle \mathbf{z}_{k}, \mathbf{r}_{k} \rangle\}$ are in correspondence with the indices of the intervals of the input $X_i$ that yielded each focal element by interval arithmetic. I.e., if interval $\mathbf{z}_{k}$ was obtained by the interval sum of $\mathbf{x}_{j_1}^1 + \dots + \mathbf{x}_{j_n}^n$ then $\mathbf{L}^{\text{aff}}(j_1, \dots, j_n) = k$. This alignment ensures that probability mass assigned to an input multi-index $\mathbf{J}$ can be pushed-forward to the output index $\mathbf{L}^{\text{aff}}(\mathbf{J})$.

\begin{proposition}[Push-forward of an imprecise multivariate copula through IBS addition]
\label{pushforward-imprecise-copula}
Let $u^*=(\gamma^1_{\ell_1},\dots,\gamma^m_{\ell_m})$ be a target grid point on the output discrete quasi-copula domain and define the index set
\[
\mathcal{I}(u^*) := \left\{\mathbf J \ \middle|\  L^{\text{aff}}_k(\mathbf J)\le \ell_k\ \text{for all }k=1,\dots,m\right\}.
\]

The value of the output imprecise copula at $u^*$ is obtained by:
\begin{align*}
\underline{Q}_{\mathbf{S}}(u^*) 
&= \max \left( \sum_{\mathbf J \in \mathcal{I}(u^*)} \underline{m}_{\mathbf{J}}, \quad 1 - \sum_{\mathbf J\notin \mathcal{I}(u^*)} \overline{m}_{\mathbf{J}} \right), \\[6pt]
\overline{Q}_{\mathbf{S}}(u^*) 
&= \min \left( \sum_{\mathbf J \in \mathcal{I}(u^*)} \overline{m}_{\mathbf{J}}, \quad 1 - \sum_{\mathbf J\notin \mathcal{I}(u^*)} \underline{m}_{\mathbf{J}} \right).
\end{align*}

These are discrete quasi-copulas defining a coherent discrete imprecise copula.
\end{proposition}

\begin{remark}
The propagation is exact at the grid nodes of the focal intervals because they align perfectly with the output focal elements via the affine map. 
This method implicitly constructs a stepwise constant function. The discrete quasi-copula envelopes could be extended to full quasi-copulas via patchwork methods \cite{kokolbukovsekExtendingMultivariateSubquasicopulas2024}. While the constructed imprecise copula is coherent in the discrete sense, the coherence of an extension on the continuous sense would need to be considered.
\end{remark}

\subsection{Activation functions}
We distinguish two cases of activation functions: first univariate, monotone, injective activations, such as sigmoids, hyperbolic tangent, leaky ReLU. Secondly, we consider the particular case of ReLU, still monotone but not injective. IBS propagation is the same in both cases:
\begin{proposition}[Monotone activation of an IBS]
\label{activation-ibs}
Let $X$ be a variable represented by a normalized IBS $I_X=\bigl\{\langle \mathbf{x}_j,[\underline m_j,\overline m_j]\rangle \mid j=1,\dots,N\bigr\}$,
where the focal sets $\mathbf{x}_j=[\underline x_j,\overline x_j]\subset\mathbb{R}$ are intervals ordered in increasing order.

Let $\sigma:\mathbb{R}\to\mathbb{R}$ be a monotone (non-decreasing) measurable
function and define $Z=\sigma(X)$. For each focal element $\mathbf{x}_j$, define its image $\mathbf{z}_j := \sigma(\mathbf{x}_j) := [\sigma(\underline x_j),\sigma(\overline x_j)]$
and construct the IBS $I_Z := \bigl\{\langle \mathbf{z}_j,[\underline m_j,\overline m_j]\rangle : j=1,\dots,N\bigr\}$ which is a sound outer approximation of the credal set under the activation.
\end{proposition}

The lack of injectivity in ReLU is a source of difficulties. The first one relates to the appearance of a mass atom in the quasi-copula envelopes below which they become ill-defined. The second relates to the identification of where this mass atom appears. In our imprecise setting, this identification is not possible, so we develop a sound way to approximate it. 

Let $\mathbf{X} = (X_1,\dots,X_n)$ be variables described by normalized IBS $I_{X_i}$ and dependent according to a coherent imprecise copula $[\underline{Q}_{\mathbf{X}},\overline{Q}_{\mathbf{X}}]$. Let $I_{Z_i}$ be the ReLUed IBSs of $I_{X_i}$ obtained by applying Proposition \ref{activation-ibs} with $Z_i = \text{ReLU}(X_i)$.

For each $i$, let $\underline{w}_i = \underline{\text{Bel}}_{X_i}((-\infty,0])$ and $\overline{w}_i = \overline{\text{Pl}}_{X_i}((-\infty,0])$ denote the lower and upper probability bounds of the event $\{X_i<0\}$ induced by the IBS $I_{X_k}$ (conservatively taken on the broadest p-box sense given by Prop. \ref{ibs-bel-pbox}). These bounds satisfy $0\le \underline{w}_i \le \overline{w}_i \le 1$ and represent the minimal and maximal probability mass that will collapse to the atom $\{0\}$ after ReLU. 

We then split the copula domain $[0,1]^n$ into two regions: $U^{+} = \Bigl\{\mathbf{u}=(u_1,\dots,u_n)\in[0,1]^n \mid u_i > \overline{w}_i\ \Bigr\}$ and $U^{-} = [0,1]^n \setminus U^{+}$. For every $\mathbf{u}\in U^{-}$, the ReLU transform may collapse a non-trivial probability mass to $\{0\}$ on at least one coordinate $k$, creating additional uncertainty about the joint allocation of that mass.

To restrict all future evaluations to the well-defined domain beyond the degenerate mass atoms, we selectively merge focal elements of the marginal IBSs which collapse to $0$ after the ReLU. This choice ensures that all quasi-copula envelope evaluations only take place in its well-defined domain, therefore implicitly becoming quasi-sub-copulas on the quantile domain above the degenerate probability atom. This approach to handling dependencies after ReLU comes at the cost of a conservative outer-approximation.

\begin{proposition}[ReLU-aware merger of IBS focal elements]
\label{relu-ibs}
Let $X$ be a real-valued random variable represented by a normalized interval
belief structure $I_X=\bigl\{\langle \mathbf{x}_j,[\underline m_j,\overline m_j]\rangle \mid j=1,\dots,N\bigr\}$. Let $\sigma=\mathrm{ReLU}$ and define $Z=\sigma(X)=\max(0,X)$. We define
\[
j_0 := \max\{j\in\{0,\dots,N\} \mid \underline{x_j} \le 0\},
\]
with the convention $j_0=0$ if $\underline{x_1}>0$.

We construct the merged focal intervals $\mathbf{z}_1 := [0,\,\max_{1\le j\le j_0}\max(0,\overline x_j)]$. For $\ell=2,\dots,N-j_0+1$, we then have:
\[
\mathbf{z}_\ell := [\underline x_{j_0+\ell-1},\,\overline x_{j_0+\ell-1}].
\]
Therefore the strictly positive focal intervals are left unchanged under ReLU. The resulting merged IBS is then given by $I_Z^{\mathrm{m}}
:= \bigl\{\langle \mathbf{z}_\ell,[\underline n_\ell,\overline n_\ell]\rangle \mid \ell=1,\dots,N-j_0+1\bigr\}$,
with mass bounds
\[
\underline n_1 := \text{Bel}_X((-\infty,0]),
\qquad
\overline n_1 := \text{Pl}_X((-\infty,0]),
\]
and for $\ell\ge 2$,
\[
\underline n_\ell := \underline m_{j_0+\ell-1},
\qquad
\overline n_\ell := \overline m_{j_0+\ell-1}.
\]

In the merged IBS, $z_1$ is the unique focal element of $I_Z^{\mathrm{m}}$ containing $0$, and it aggregates all mass that may be mapped to $0$ by ReLU (i.e., originating from values $\le 0$ before activation). Moreover, $I_Z^{\mathrm{m}}$ is a sound outer approximation of applying ReLU to the credal set defined by the original IBS.
\end{proposition}

After merging, all quantile grid points satisfy $u_i \ge \overline{w}_i$ or $u_i=0$ for the edges, where the quasi-copulas are $0$ by definition. Therefore once the merger has taken place, the imprecise copula can be evaluated and propagated through the activation function. In the case of injective activations, no merger is required.

Let $\sigma_i:\mathbb{R}\to\mathbb{R}$ be measurable activation functions and define $Z_i=\sigma_i(X_i)$, $Z=(Z_1,\dots,Z_n)$. For each marginal $i$, construct an output IBS $I_{Z_i}$ by applying Prop. \ref{activation-ibs} and the ReLU-aware merger if the activation in question is ReLU (Prop. \ref{relu-ibs}). Let $\left \{ \{\gamma^i_{j_i}\}_{j_i=1}^{M_i} \right \}_{i=1}^n$ be the resulting quantile grid for $Z_i$. There exists a memory index map $\textbf L^{\text{act}}:\prod_i\{1,\dots,N_i\}\to\prod_i\{1,\dots,M_i\}$, constructed analogously to $\textbf L^{\text{aff}}$, such that each $R_{\mathbf J}$ maps into a unique output cell.

\begin{proposition}[Push-forward of an imprecise copula through an activation]
\label{push-forward-activation}
For any point in the output quantile grid $u^*=(\gamma^1_{\ell_1},\dots,\gamma^n_{\ell_n})$ define
\[
\mathcal{I}(u^*) := \left\{\mathbf J \ \middle|\  L^{\text{act}}_i(\mathbf J)\le \ell_i\ \text{for all }i=1,\dots,n\right\}.
\]
We define the push-forward copula envelopes
\[
\underline Q_Z(u^*)
:=\max\!\left\{
\sum_{\mathbf J \in \mathcal{I}(u^*)}\underline m_{\mathbf J},\;
1-\sum_{\mathbf J\notin \mathcal{I}(u^*)}\overline m_{\mathbf J}
\right\},
\]
\[
\overline Q_Z(u^*)
:=\min\!\left\{
\sum_{\mathbf J \in \mathcal{I}(u^*)}\overline m_{\mathbf J},\;
1-\sum_{\mathbf J\notin \mathcal{I}(u^*)}\underline m_{\mathbf J}
\right\}.
\]

Then $[\underline Q_Z,\overline Q_Z]$ defines a coherent discrete imprecise
copula for $Z$ on the output quantile grid.  
\end{proposition}

\subsection{Full propagation algorithm}
We can now compose the affine and activation operations on coupled IBSs to propagate them through a full feed-forward neural network, in Algorithm \ref{alg:coupled-dsi-copula-nn-concise}.

\begin{algorithm}[t]
\caption{Coupled IBS propagation through a neural network}
\label{alg:coupled-dsi-copula-nn-concise}
\begin{algorithmic}[1]
\Require Input p-boxes $\{[\underline F_i,\overline F_i]\}_{i=1}^{h_0}$; grids $\{\alpha^{i,0}_j\}_{j=0}^{N_{i,0}}$;
input imprecise copula $C^0 = [\underline{Q}_{\mathbf X^0}, \overline{Q}_{\mathbf X^0}]$; network $\{(W^k,b^k,\sigma^k)\}_{k=0}^{L-1}$.
\Ensure $\{d^L_{X^L_r}\}_{r=1}^{h_L}$ and imprecise copula $[\underline{Q}_{L}, \overline{Q}_{L}]$ on $\mathcal U^L=\prod_{r=1}^{h_L}\{\alpha^{r,L}_j\}$.

\Statex \textbf{Init:} $(d^0_{X^0_i},\{\alpha^{i,0}_j\})\gets\textsc{PboxToIBS}([\underline F_i,\overline F_i],\{\alpha^{i,0}_j\})$ for $i=1,\dots,h_0$;
$C^0(\mathbf u)\gets C_{\mathbf X^0}(\mathbf u)$ for all $\mathbf u\in\mathcal U^0$.

\For{$k=0$ to $L-1$}
  \Statex \textbf{Affine:} $S^{k+1}=W^kX^k+b^k$
  \State $(\{d^{k+1}_{S^{k+1}_r}\},\{\gamma^{r,k+1}\},L^{k,\mathrm{aff}})\gets \textsc{AffineStep}(\{d^k_{X^k_i}\},\{\alpha^{i,k}\},C^k,W^k,b^k)$
  \Comment{Prop. \ref{sum-ibs}}
  \State $C^{k+1,\mathrm{aff}}\gets \textsc{PushFwdCopula}(C^k,\{\alpha^{i,k}\},\{\gamma^{r,k+1}\},L^{k,\mathrm{aff}})$
  \Comment{Prop. \ref{pushforward-imprecise-copula}}

  \Statex \textbf{Activation:} $X^{k+1}=\sigma^k(S^{k+1})$
  \State $(\{d^{k+1}_{X^{k+1}_r}\})\gets \textsc{ActIBS}(\{d^{k+1}_{S^{k+1}_r}\},\sigma^k)$
  \Comment{Prop. \ref{activation-ibs}}
  \State $(\{d^{k+1}_{X^{k+1}_r}\},\{ \alpha^{r,k+1} \},L^{k,\mathrm{act}})\gets \textsc{Merge}(\{d^{k+1}_{X^{k+1}_r}\},\sigma^k)$
  \Comment{Prop. \ref{relu-ibs}}
  \State $C^{k+1}\gets \textsc{PushFwdCopula}(C^{k+1,\mathrm{aff}},\{\gamma^{r,k+1}\},\{\alpha^{r,k+1}\},L^{k,\mathrm{act}})$
  \Comment{Prop. \ref{push-forward-activation}}
\EndFor
\State \Return $\big(\{d^L_{X^L_r}\}_{r=1}^{h_L},\, C^L\big)$
\end{algorithmic}
\end{algorithm}

The algorithm has exponential complexity first with respect to the amount of neurons per layer which leads to an exponential explosion of the copula quantile grid domain; and with respect to the number of focal elements in the IBS marginals, which also grow exponentially after each affine step.
\subsection{Quantitative verification}
The propagation of coupled IBSs through a neural network provides the reachable states with sound probabilistic information and a rich description of the domain of the outputs of the neural networks and their dependencies.

The resulting structure allows us, in particular, to verify probabilistic properties. Given a linear safety property $Hy \leq w$ on the network output vector $y$ we can compute sound bounds for the probability of the property being satisfied. Practically, an extra affine step on the coupled IBS is used to evaluate $z = Hy - w$. The probability bounds of the linear property of interest are then given by the belief and plausibility (Prop. \ref{ibs-bel-pbox}) of the $z$ IBS evaluated at $0$. The copula volumes also allow us to verify a conjunction of linear properties.

\section{Conclusions}
We have developed a pipeline to propagate imprecisely coupled interval belief structures through affine transforms and activation layers. While these operations may be of interest in other fields, we use them for neural networks reachability analysis, obtaining imprecise descriptions of the network's outputs and their dependence, with application to quantitative verification. 

Future work includes a computational implementation in order to evaluate this methodology on well-established benchmarks for probabilistic verification. This may require developing more careful heuristic merger strategies through the propagation to keep computation tractable.
Difficulties may also arise in modelling features and copulas in a fully imprecise manner. 
Finally, the probabilistic information obtained after the propagation is much more detailed than needed for the verification of linear safety properties. This paves the way for further interpretability applications on the imprecise multivariate distributions obtained such as robustness certification or sensitivity analysis.

\begin{credits}
\subsubsection{\ackname} This work was partially supported by
the SAIF project, funded by the “France 2030” government
investment plan managed by the French National Research
Agency, under the reference ANR-23-PEIA-0006.

\subsubsection{\discintname}
The authors have no competing interests. 
\end{credits}
%
%
\bibliographystyle{splncs04}
\bibliography{refs}

\end{document}